\pdfoutput=1

\documentclass[11pt]{article}

\usepackage{acl}

\usepackage{times}
\usepackage{latexsym}

\usepackage{tcolorbox}

\usepackage[T1]{fontenc}

\usepackage[utf8]{inputenc}

\usepackage{color}
\usepackage{tabularx}
\usepackage{tcolorbox}
\usepackage{xcolor}
\usepackage{colortbl}
\usepackage{tabu}
\usepackage{booktabs}
\usepackage{multirow}

\usepackage{graphicx} 
\usepackage{float}
\usepackage{subfigure} 
\usepackage{amssymb}

\usepackage{hyperref}
\usepackage{tablefootnote}
\usepackage{xspace}
\usepackage[normalem]{ulem}

\usepackage{float}
\usepackage{amsmath}
\usepackage{bm}
\usepackage{algorithmicx}
\usepackage{algorithm}
\usepackage{algpseudocode}

\usepackage{arydshln}

\usepackage{amssymb}
\usepackage{pifont}
\usepackage{enumitem}
\usepackage{fontawesome}  
\usepackage{xspace}       
\usepackage{hyperref}

\usepackage{microtype}

\newcommand{\ours}{\textsc{SuCEA}\xspace}
\newcommand{\fm}{\textsc{FoolMeTwice}\xspace}
\newcommand{\wice}{\textsc{Wice}\xspace}
\newcommand{\base}{\textsc{RALM}\xspace}
\newcommand{\claimdecompose}{\textsc{ClaimDecomp}\xspace}
\newcommand{\qabrief}{\textsc{QA Briefs}\xspace}
\newcommand{\programfc}{\textsc{ProgramFC}\xspace}
\newcommand{\minicheck}{\textsc{MiniCheck}\xspace}

\newcommand{\eg}{\hbox{\emph{e.g.,}}\xspace}
\newcommand{\ie}{\hbox{\emph{i.e.}}\xspace}

\title{\ours: Reasoning-Intensive Retrieval for Adversarial Fact-checking through Claim Decomposition and Editing}

\author{
  Hongjun Liu\textsuperscript{1,2} \quad 
  Yilun Zhao\textsuperscript{3} \quad 
  Arman Cohan\textsuperscript{3,4} \quad Chen Zhao\textsuperscript{1, 2}\\
  \textsuperscript{1}New York University \quad
  \textsuperscript{2}NYU Shanghai \quad
  \textsuperscript{3}Yale University \quad\quad
  \textsuperscript{4}Allen Institute for AI
}

\begin{document}
\maketitle

\begin{minipage}[t]{2\linewidth}
\vspace{-1.75cm}
  \centering
  \href{https://github.com/HolaYan/SUCEA}{{\faGithub{}}\xspace\texttt{https://github.com/HolaYan/SUCEA}} 
\vspace{0.5cm}
\end{minipage}

\begin{abstract}
Automatic fact-checking has recently received more attention as a means of combating misinformation. Despite significant advancements, fact-checking systems based on retrieval-augmented language models still struggle to tackle adversarial claims, which are intentionally designed by humans to challenge fact-checking systems.
To address these challenges, we propose a training-free method designed to rephrase the original claim, making it easier to locate supporting evidence. 
Our modular framework, \ours, decomposes the task into three steps: 1) \emph{Claim Segmentation and Decontextualization} that segments adversarial claims into independent sub-claims; 2) \emph{Iterative Evidence Retrieval and Claim Editing} that iteratively retrieves evidence and edits the subclaim based on the retrieved evidence; 3) \emph{Evidence Aggregation and Label Prediction} that aggregates all retrieved evidence and predicts the entailment label.
Experiments on two challenging fact-checking datasets demonstrate that our framework significantly improves on both retrieval and entailment label accuracy, outperforming four strong claim-decomposition-based baselines.

\end{abstract}

\begin{figure}[!t]
    \centering
    \includegraphics[width = \linewidth]{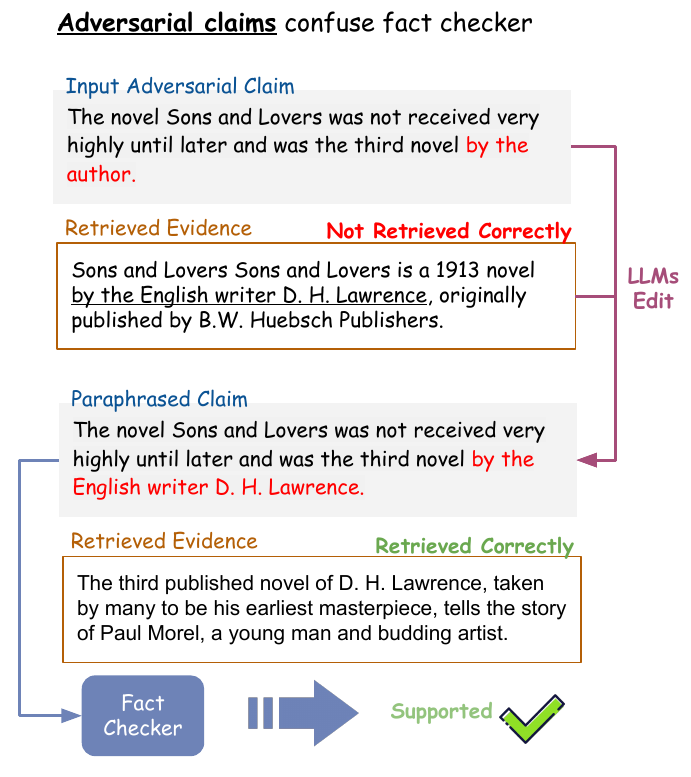}
    \caption{Examples of challenges in fact-checking adversarial claims. 
    These claims are deliberately crafted by humans using strategies to deceive fact-checking systems. 
    Even state-of-the-art approaches (\ie RALM) face significant difficulties in accurately verifying these claims. 
    }
    \label{fig:task}
\end{figure}
\section{Introduction}
In the era of information overload, news items are often rapidly and indiscriminately published by news outlets, which can lead to the spread of misinformation~\cite{chen2023combating}. Consequently, it is essential to automatically verify these claims that circulate online~\cite{thorne-etal-2018-fever, min-etal-2023-factscore, akhtar-etal-2023-multimodal}.
The current research landscape in automated fact-checking mainly adopts a retrieval-augmented language model (RALM) framework~\cite{asai2024reliable, gao2024ralmsurvey}. It involves a separate information retrieval system, such as term-matching methods such as BM25 or dense retrievers~\cite{karpukhin2020dense}, to locate evidence from extensive text collections used as an external datastore. 
The retrieved output is then fed to a large language model (LLM) along with the input claim to predict the entailment labels. With the advancements of LLMs~\cite{touvron2023llama, OpenAI2023GPT4TR, groeneveld-etal-2024-olmo}, these automated fact-checking systems have shown remarkable performance improvements across various fact-checking benchmark datasets~\cite{fever-2023-fact, kamoi-etal-2023-wice,zhao2024findverexplainableclaimverification}.

Nevertheless, existing RALM approaches for fact-checking still lack robustness, especially when dealing with claims adversarially crafted by humans~\cite{eisenschlos-etal-2021-fool, kamoi-etal-2023-wice}. Human claim writers often deliberately reduce the textual overlap between claims and evidence, \eg
omit crucial information (in \autoref{fig:task}, the author name \underline{D. H. Lawrence} is absent). Therefore, it requires in-depth reasoning to identify relevant evidence that goes beyond surface form matching, \ie reasoning-intensive retrieval~\cite{su2024bright,zhao2024financemathknowledgeintensivemathreasoning}.
This challenge is further exacerbated by the complexity of such claims, which frequently require verification of multiple components (\eg the claim in \autoref{fig:task} involves checking two distinct elements).

To address the challenge of adversarial fact-checking, we introduce \textbf{\ours}, a \underline{\textbf{\textsc{Su}}}b\underline{\textbf{C}}laim-based \underline{\textbf{E}}vidence-\underline{\textbf{A}}ugmented Enhancement Fact-checking framework without additional LLM fine-tuning.
\ours contains three key modules: 
(1) \textbf{Claim Segmentation and Decontextualization}, which leverages LLMs to segment and decontextualize the input claims into several \textit{atomic} segments.
(2) \textbf{Iterative Evidence Retrieval and Claim Editing}, which iteratively retrieves evidence and  rewrites each sub-claim that is lacking sufficient support.
(3) \textbf{Evidence Aggregation and Label Prediction}, which aggregates all retrieved evidence for final entailment label prediction. 

Compared with existing fact-checking systems, \ours features the following advancements: 
(1) Our claim editing module tackles reasoning-intensive retrieval challenge through  
transforming these adversarially designed segments into more retriever-friendly ones (\eg by increasing textual overlap, simplifying uncommon phrases, or adding key information). 
(2) Editing claims from scratch is non-trivial, as it is unclear where evidence retrieval struggles. Our approach addresses this by grounding the editing process in retrieved evidence, which provides essential guidance, such as relevant topics and person names.
(3) Our decomposition strategy further simplifies reasoning challenges by breaking claims into decontextualized segments and increasing the lexical overlap between decomposed sub-claims and evidence.

On \fm, \ours demonstrates substantial improvements over RALM fact-checking systems, with 7.5\% boost in fact-checking accuracy.
This enhancement is primarily attributed to improved evidence retrieval, reflected in an 11.0\% increase in retrieval Recall@10 under \texttt{Llama-3.1-70B}. Additionally, \ours outperforms baselines on another real-world complex fact-checking dataset, \wice, demonstrating the general applicability of our framework for verifying complex claims. We further conduct a detailed ablation study and qualitative analysis, showing that the claim editing module simplifies and enhances evidence retrieval, while the claim segmentation module reduces distractions during retrieval.

\section{The Real-world Adversarial Fact Checking Task}
This section first defines the task setup (\S\ref{sec:problem}) and then discusses the datasets used in our study (\S\ref{sec:dataset}). 

\subsection{Problem Formulation}
\label{sec:problem}

We formulize the task of adversarial fact-checking in open-domain settings as follows: 
The input consists of a claim \( C \) and a large text collection $D$, which serves as an external datastore. The task is to first retrieve relevant evidence $e_1, ..., e_k$, with the retrieval performance evaluated by the accuracy and recall of gold evidence (\S\ref{sec:exp_setup}). Following this, the fact-checking system assigns one of entailment labels $L_C$ to the claim: \{\emph{supported}, \emph{refutes}, \emph{not enough evidence}\}.

\begin{table}[t!]
\centering 
\small
\resizebox{\columnwidth}{!}{
\renewcommand{\arraystretch}{1.2}
\begin{tabular}{lrr}
\toprule
\textbf{Property}                            & \textbf{\fm} & \textbf{\wice}  \\
\midrule
\multicolumn{3}{c}{\textbf{Knowledge Corpus}} \\
Source & Wikipedia & Wikipedia \\
Knowledge Corpus Size &21,015,325 &21,015,325\\
\noalign{\vskip 0.5ex}\hdashline\noalign{\vskip 0.5ex}
\multicolumn{3}{c}{\textbf{Evidence Retrieval \& Fact Checking}} \\
\# Evidence (avg.)    & 1.29 & 3.94 \\
Claim Token Length (avg.)    &  14.03  & 24.20\\

Test Set Size (\# Claims)   & 200  &  358   \\
\quad \# Entailed Claims &103 &111\\
\quad \# Refuted Claims &97 &32\\
\quad \# Partially Supported Claims & N/A &215\\
\toprule
\end{tabular}
}
\caption{Basic statistics of the \fm and \wice test sets used in our experiments. Due to limited budgets, we evaluate on the full test set of \wice (358 instances) and 200 random samples from the \fm test set.}
\label{tab:basic-stats}
\end{table}

\subsection{Datasets}
\label{sec:dataset}
We mainly conduct experiments and analysis on the adversarial dataset \fm~\cite{eisenschlos2021fool}. 
We also choose \wice~\cite{kamoi2023wicerealworldentailmentclaims}, a long-form fact-checking dataset, to demonstrate the broader applicability of \ours.
\autoref{tab:basic-stats} presents an overview of the test set statistics used in this study.

\begin{itemize}[leftmargin=*]
    \itemsep0em 
    \item \textbf{\fm} is a challenging fact-checking dataset with adversarially written claims. The dataset is collected through a multi-player game, where human players are allowed to modify claims to more challenging versions to deceive other players (\ie unable to verify). Through this process, the collected claims are more complex and often require a higher order of reasoning (\eg inference about time, understanding hyponymy, phrase paraphrasing, etc). 

    \item \textbf{\wice} focuses on real-world claim entailment grounded on Wikipedia pages. It is equipped with long-form claims with numerous pieces of statements. These long-form claims create challenges for fact-checking systems, especially evidence retrieval components. As it requires retrievers to find \textit{complete} set of evidence. 
    In this study, we expand the retrieval process of the \wice dataset to open-domain settings. 
\end{itemize}

\begin{figure}[!t]
    \centering
    \includegraphics[width = 1.0\linewidth]{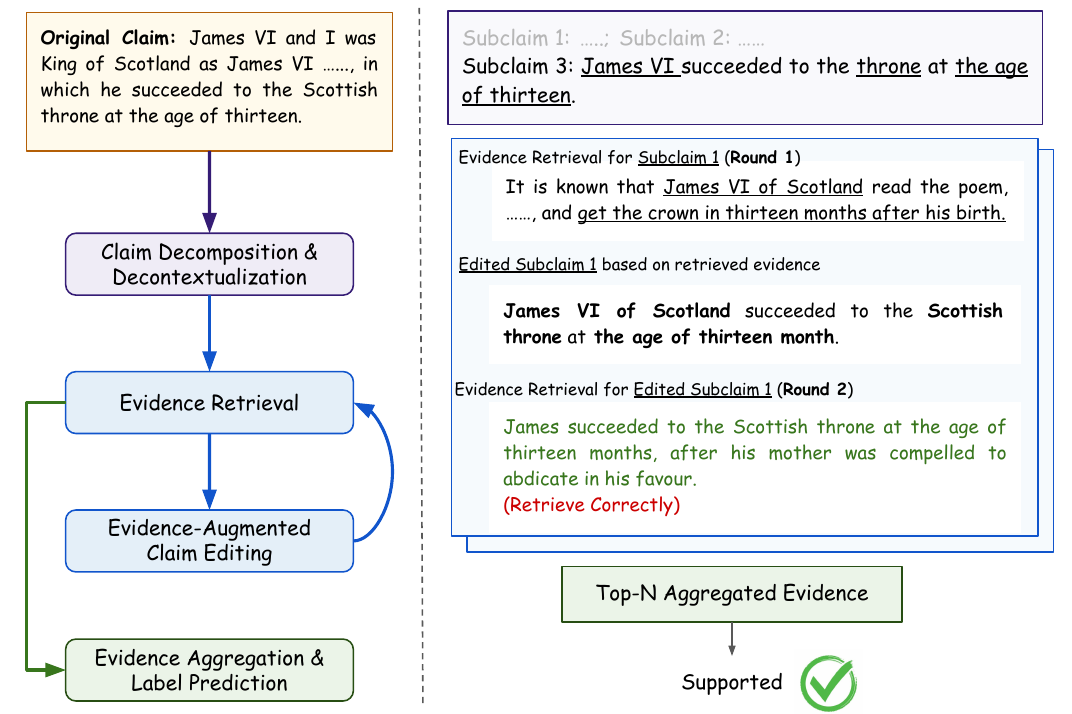}
    \caption{An illustration of \ours pipeline (left) and its workflow for adversarial fact-checking using an example from the \fm dataset (right). The input claim is first segmented and decontextualized by LLMs. For each sub-claim, a first round of evidence retrieval is conducted, followed by sub-claim revision and a second round of evidence retrieval. The final step is label prediction for the claim. 
    }
    \label{fig:model}
\end{figure}

\section{Method}
In this section, we present our proposed framework, \ours, for adversarial fact-checking. 
This task presents two significant challenges for existing RALM-based fact-checking systems: the relevant knowledge cannot be directly retrieved through lexical or semantic matching alone, but instead require intensive reasoning; the complexity of the input claims require verifying multiple sub-components and aggregating the results.

More specifically, \ours contains three key modules (Figure~\ref{fig:model}): 
(1) The \textbf{Claim Segmentation and Decontextualization} module breaks the input claims into multiple sub-claims, and then rewrites each sub-claim to make it context-free and independent (\S\ref{sec:model_sts}).
(2) The \textbf{Iterative Evidence Retrieval and Claim Editing} module retrieves evidence for each sub-claim.
When retrieval fails, the LLMs paraphrase the sub-claims based on the retrieved evidence by using our specialized constraint prompt and then initiate a new round of retrieval (\S\ref{sec:model_ce}). 
(3) The \textbf{Evidence Aggregation and Label Prediction} module finally combines retrieved evidence and makes the final entailment label prediction (\S\ref{sec:model_lp}).

\subsection{Claim Segmentation and Decontextualization}
\label{sec:model_sts}
Real-world claims often contain multiple facts that need to be verified. For example, verifying the input claim in Figure~\ref{fig:model} entails identifying who James VI is, exploring his biography, and determining his age at the time he ascended the Scottish throne.
Verifying all sub-claims simultaneously is challenging for current RALM systems, mainly because the retrieval component cannot find all evidence pieces at once. Often, the top-ranked evidence may address only a few fact units. To tackle this issue, we follow existing work~\cite{min2023factscorefinegrainedatomicevaluation} that instructs LLMs to decompose the claims into decontextualized segments. 
Each segment represents an atomic fact, which can then be fact-checked independently. We detail this module as follows:

\paragraph{Claim Segmentation.} Given an input claim $C$, we prompt LLMs to decompose into a sequence of sub-claims $C_1, C_2,..., C_n$. 
We include the prompt details in \autoref{fig:seg} in the appendix.

\paragraph{Claim Decontextualization.} For each sub-claim $C_i$, we further prompt LLMs to ensure it is a standalone statement by adding entity name or rewriting pronouns~\cite{gunjal2024molecular}.  We include the prompt details in \autoref{fig:decontext} in appendix.

\subsection{Iterative Evidence Retrieval and Claim Editing}
\label{sec:model_ce}
A key challenge in fact-checking adversarial claims is that there is less textual overlap between claim and evidence. 
For example, writers paraphrase the evidence \emph{``Sister Carrie was also criticized for \underline{never mentioning the name of God}.''} into the statement \emph{``Sister Carrie was criticized for taking the \underline{Lord’s title in vain}''}. 
Therefore, retrievers need in-depth reasoning to identify relevant evidence that goes beyond surface form matching (\eg multi-hop reasoning by adding the missing information as intermediate hop). 
To this end, we propose a reverse engineering approach that prompts LLMs to paraphrase these sub-claims back to their original form by eliminating the misleading controversial content and adding missing or modified key information. 
However, directly prompting LLMs to paraphrase back sub-claims is non-trivial, as LLMs have no explicit direction. 
We instead first retrieve evidence for original sub-claims and mainly use it as a hint to guide LLMs towards right editing directions (\eg \emph{``\underline{Lord’s title in vain}''} $\rightarrow$ \emph{``\underline{never mentioning the name of God}''}). We detail this module as follows.

\paragraph{First-Round Evidence Retrieval.}
For each sub-claim $C_i$, we adopt the off-the-shelf retrievers (\eg TFIDF or Dense Retrieval) to find top-$k$ evidence from text collections (\ie Wikipedia). Note that these claims are adversarially written to fool the retrieval systems, thus we do not expect to find evidence in one shot, but instead use such information to guide claim editing that we present next.

\paragraph{Sub-claim Paraphrase.}
For each sub-claim $C_i$ and corresponding top-$k$ evidence $e_1, ..., e_k$, we prompt LLMs to paraphrase $C_i$ to $\hat{C_i}$ so that it becomes retriever-friendly. Note that we directly add the constraints into prompts so that it only adds information from the evidence, rather than from LLM's parametric knowledge (which is more likely to hallucinate). 
Additionally, to ensure high generation quality, we enhance the process by applying specific paraphrasing guidelines, such as completing missing named entities, numerical values, and locations, correcting counterfactual errors, and utilizing a one-shot approach.
\autoref{fig:paraphrase_prompt} in Appendix presents the prompt details.

\paragraph{Second-Round Evidence Retrieval.} For each paraphrased sub-claim $\hat{C_i}$, we adopt the off-the-shelf retrievers again to find top-$k$ evidence.
The paraphrasing of sub-claims is intended to be retrieval-friendly, thereby improving the likelihood of retrieving relevant evidence.

\subsection{Evidence Aggregation and Label Prediction}
\label{sec:model_lp}
This module aims to aggregate the retrieved evidence for each sub-claim and utilize the combined information to predict the final entailment label.

\paragraph{Aggregate Evidence.} Given $n$ sub-claims, we retrieve a total of $k \times n$ evidence pieces. We then prompt the LLMs to rerank these pieces and select the top-$k$ most relevant ones. \autoref{fig:ranker} in Appendix presents the prompt details.

\paragraph{Entailment Label Prediction.} We next prompt LLMs to predict the final entailment label
based on the claim $C$ and its top-$k$ retrieved supporting evidence as context. \autoref{fig:fm2_entail} in Appendix presents the  prompt details.

\section{Experiment}

We examine our framework for the adversarial fact-checking task, using two benchmark datasets, \fm and \wice \ (discussed in \S\ref{sec:dataset}).

\subsection{Experiment Setup}\label{sec:exp_setup}

\paragraph{\ours Implementation Details.}
We use \texttt{GPT-4o-mini} \footnote{\texttt{gpt-4o-mini-2024-07-18}}\cite{OpenAI2023GPT4TR} and \texttt{Llama-3.1-70B} \cite{dubey2024llama3} as the backbone LLMs for segmentation and sub-claim editing. We further compare with other open-source LLMs, including  \texttt{Llama-3.1-8B} \cite{dubey2024llama3}, \texttt{Mistral 9B} \cite{mistral2023moe}, and \texttt{Gemma-2-9B} \cite{gemma2}. We present the results in appendix (\autoref{tab:smallmodelresult}). We adopt three different retrievers: 
 \texttt{TF-IDF}~\cite{schutze2008introduction} that focuses on lexical similarity, and  \texttt{Contriever} \cite{izacard2022unsuperviseddenseinformationretrieval} that encodes both claims and passages into embeddings and compares the vector distance as relevance measure. We follow \texttt{Contriever} setup and use the Dec. 20, 2018 version of Wikipedia as the retrieval corpus. We preprocess the Wikipedia so that each passage contains one hundred words.

\paragraph{Baselines.} We compare \ours with the following  baselines: (1) \emph{Retrieval-augmented Generation} that finds top-k evidence from Wikipedia, and then uses LLMs to predict entailment labels;
(2) \emph{Sub-claim Generation}: \claimdecompose \cite{chen2022generatingliteralimpliedsubquestions}, \qabrief \cite{fan2020generatingfactcheckingbriefs} and \minicheck \cite{tang-etal-2024-minicheck} decompose claims into a set of sub-claims and aggregate the results of verifying the sub-claims.
(2) \emph{Reasoning Program Generation}: \programfc \cite{pan2023factcheckingcomplexclaimsprogramguided}  decomposes a fact-checking task into a sequence of sub-tasks through a program, executes each task and aggregates results to verify the claim.

\paragraph{Metrics.} 
We use \textbf{Accuracy} to measure the alignment between predicted entailment labels and the ground truth for fact checking.
For \emph{evidence retrieval}, we employ two widely-used metrics: 
\textbf{Retrieval Accuracy (\emph{RAcc.})}, which assesses whether at least one relevant evidence piece is in the top-$k$ results (following \fm); 
and \textbf{Retrieval Recall@\bm{$k$} (\emph{Recall@\bm{$k$}})}, which measures the proportion of relevant evidence pieces retrieved among the top-$k$ results. 

\begin{table}[t]
\centering
\small
\resizebox{\columnwidth}{!}{
\renewcommand{\arraystretch}{1.2}
\begin{tabular}{lcccc}
\toprule
\multirow{2}{*}{System} & \multicolumn{2}{c}{\fm} & \multicolumn{2}{c}{\wice} \\
\cmidrule{2-5}
& Contriever & TFIDF & Contriever & TFIDF \\
\midrule
\multicolumn{5}{l}{\textbf{GPT-4o-mini}} \\
\quad \base & 67.5 & 61.5 & 33.7 & 27.3 \\
\quad \claimdecompose & 69.5 (\textcolor{red}{+2.0})  & 63.0 (\textcolor{red}{+1.5}) & 35.2 (\textcolor{red}{+1.5}) & 31.0 (\textcolor{red}{+3.7}) \\
\quad \qabrief & 67.5(\textcolor{red}{+0.0})	&62.5(\textcolor{red}{+1.0}) & 36.2(\textcolor{red}{+2.5}) &29.7(\textcolor{red}{+2.4}) \\
\quad \programfc & 69.0 (\textcolor{red}{+1.5}) & 63.0 (\textcolor{red}{+1.5}) & 38.4 (\textcolor{red}{+4.7}) & 29.9 (\textcolor{red}{+2.6}) \\
\quad \minicheck & 70.0 (\textcolor{red}{+2.5}) & 64.0 (\textcolor{red}{+2.5}) & 33.7 (\textcolor{red}{+0.0}) & 30.8 (\textcolor{red}{+3.5})\\

\noalign{\vskip 0.5ex}\hdashline\noalign{\vskip 0.5ex}

\quad \textbf{\ours} & \textbf{75.0 (\textcolor{red}{+7.5})} & \textbf{68.5 (\textcolor{red}{+7.0})} & \textbf{39.0 (\textcolor{red}{+5.3})} & \textbf{33.2 (\textcolor{red}{+5.9})} \\
\midrule
\multicolumn{5}{l}{\textbf{LLama-3.1-70B}} \\
\quad \base & 65.5 & 61.5 & 31.2 & 25.7 \\
\quad \claimdecompose & 67.5 (\textcolor{red}{+2.0}) & 64.0 (\textcolor{red}{+2.5}) & 33.0 (\textcolor{red}{+1.8}) & 27.1 (\textcolor{red}{+1.4}) \\
\quad \qabrief & 66.0(\textcolor{red}{+0.5})	&64.0(\textcolor{red}{+2.5}) & 33.5(\textcolor{red}{+2.3}) &28.2(\textcolor{red}{+2.5}) \\
\quad \programfc & 66.0 (\textcolor{red}{+0.5}) & 64.5 (\textcolor{red}{+3.0}) & 33.9 (\textcolor{red}{+2.7}) & 29.3 (\textcolor{red}{+3.6}) \\
\quad \minicheck & 70.5 (\textcolor{red}{+5.0}) & 64.0 (\textcolor{red}{+2.5}) & 35.2 (\textcolor{red}{+4.0}) & 30.1 (\textcolor{red}{+4.4})\\

\noalign{\vskip 0.5ex}\hdashline\noalign{\vskip 0.5ex}

\quad \textbf{\ours} & \textbf{73.5 (\textcolor{red}{+8.0})} & \textbf{69.0 (\textcolor{red}{+7.5})} & \textbf{38.7 (\textcolor{red}{+7.5})} & \textbf{32.9 (\textcolor{red}{+7.2})} \\
\bottomrule
\end{tabular}
}
\caption{Fact-checking evaluation results under baselines and \ours for \fm and \wice models.
Numbers in red parentheses indicate performance improvements over the baseline (\base). }
\label{tab:factcheckresults}
\end{table}

\subsection{Main Experimental Results}


\begin{table*}[!t]
\centering
\resizebox{\textwidth}{!}{%
\addtolength{\tabcolsep}{-0.35em}
\begin{tabular}{lcccccccccccccc}
\toprule
 \multirow{4}{*}{System}  & \multicolumn{4}{c}{Top@3} & \multicolumn{4}{c}{Top@5} & \multicolumn{4}{c}{Top@10} \\
        \cmidrule(lr){2-5} \cmidrule(lr){6-9} \cmidrule(lr){10-13}
         & \multicolumn{2}{c}{Contriever} & \multicolumn{2}{c}{TFIDF} & \multicolumn{2}{c}{Contriever} & \multicolumn{2}{c}{TFIDF} & \multicolumn{2}{c}{Contriever} & \multicolumn{2}{c}{TFIDF} \\
        \cmidrule(lr){2-3} \cmidrule(lr){4-5} \cmidrule(lr){6-7} \cmidrule(lr){8-9} \cmidrule(lr){10-11} \cmidrule(lr){12-13}
        & RAcc. & Recall & RAcc. & Recall & RAcc. & Recall & RAcc. & Recall & RAcc. & Recall & RAcc. & Recall \\
\midrule

\textbf{\fm} \\\noalign{\vskip 0.5ex}

{Retriever only} & 28.5 & 24.8 & 11.0 & 8.9 & 36.0 & 31.0 & 15.5 & 12.8 & 47.5 & 42.6 & 23.5 & 19.0 \\
\ours (Llama-3.1-70B) & 40.5 ({\color{red}+12.0}) & 35.7 ({\color{red}+10.9}) & 27.5 ({\color{red}+16.5}) & 23.3 ({\color{red}+14.4}) & 47.5 ({\color{red}+11.5}) & 40.7 ({\color{red}+9.7}) & 29.5 ({\color{red}+14.0}) & 26.4 ({\color{red}+13.6}) & 51.0 ({\color{red}+3.5}) & 46.5 ({\color{red}+3.9}) & 34.5 ({\color{red}+11.0}) & 29.5 ({\color{red}+10.5}) \\
\ours (GPT-4o-mini) & 39.5 ({\color{red}+11.0}) & 34.9 ({\color{red}+10.1}) & 20.5 ({\color{red}+9.5}) & 17.4 ({\color{red}+8.5}) & 46.5 ({\color{red}+10.5}) & 41.1 ({\color{red}+10.1}) & 28.5 ({\color{red}+13.0}) & 24.0 ({\color{red}+11.2}) & 54.0 ({\color{red}+6.5}) & 49.2 ({\color{red}+6.6}) & 33.5 ({\color{red}+10.0}) & 27.9 ({\color{red}+8.9}) \\

\midrule

\textbf{\wice} \\
\noalign{\vskip 0.5ex}

{Retriever only}  & 7.0 & 3.0 & 5.3 & 2.5 & 8.7 & 3.8 & 7.0 & 3.3 & 11.7 & 5.0 & 10.9 & 4.3 \\
\ours (Llama-3.1-70B) & 11.1 ({\color{red}+4.1}) & 4.5 ({\color{red}+1.5}) & 7.8 ({\color{red}+2.5}) & 3.5 ({\color{red}+1.0}) & 14.1 ({\color{red}+5.4}) & 5.0 ({\color{red}+1.2}) & 9.5 ({\color{red}+2.5}) & 4.0 ({\color{red}+0.7}) & 16.4 ({\color{red}+4.7}) & 6.5 ({\color{red}+1.5}) & 14.0 ({\color{red}+3.1}) & 5.8 ({\color{red}+1.5}) \\
\ours (GPT-4o-mini) & 8.4 ({\color{red}+1.4}) & 3.7 ({\color{red}+0.7}) & 7.3 ({\color{red}+2.0}) & 3.1 ({\color{red}+0.6}) & 10.6 ({\color{red}+1.9}) & 4.4 ({\color{red}+0.6}) & 9.2 ({\color{red}+2.2}) & 3.7 ({\color{red}+0.4}) & 16.0 ({\color{red}+4.3}) & 6.4 ({\color{red}+1.4}) & 13.4 ({\color{red}+2.5}) & 5.7 ({\color{red}+1.4}) \\

\bottomrule
\end{tabular}
}
\caption{Comparison of retrieval performance between baseline retriever and \ours on \fm and \wice test sets under Top-k evidence (k=3,5,10). We report both Retrieval Accuracy (\textbf{RAcc.}) and Recall metrics. Numbers in red parentheses indicate absolute improvements. 
}
\label{tab:retriveresult}
\end{table*}

\paragraph{\ours outperforms baselines in adversarial claims.}
%
According to \autoref{tab:factcheckresults},\footnote{We present full results in Appendix (\autoref{tab:whole_factcheckresults}).} \ours significantly outperforms all baselines in \fm dataset. Specifically, compared with \base, the fact-checking accuracy improves from 65.5\% to 73.5\%.\footnote{Specifically, for supported claims, the result improves from 52.4\% to 64.1\%. And for refuted claims, the result improves from 79.4\% to 88.7\%.}
In contrast, baselines struggle to effectively address adversarial scenarios.  Unlike \ours, these baseline methods mainly focus on variants of claim decomposition, but do not incorporate claim editing, therefore retrievers still struggle to find relevant evidence.


\paragraph{Higher retrieval accuracy leads to better fact-checking Accuracy.}
According to \autoref{tab:retriveresult}, 
\ours greatly boosts the retrieval accuracy on all measures, which leads to an increase in fact-checking accuracy. Notably, for TFIDF-based retrieval, which is particularly susceptible to adversarial queries, we observe a 11.0\% in $RAcc.$, leading to a 7.5\% increase in fact-checking accuracy with \texttt{Llama-3.1-70B}.\footnote{It's noted that despite significance, the degree of improvement in fact-checking is smaller than retrieval. We hypothesize it is because in some instances, LLMs, especially with large parameter sizes, use their parametric knowledge rather than retrieved evidence for entailment label prediction. } 


\paragraph{Dense retrieval significantly outperforms TFIDF.} 
Specifically, Contriever achieves 73.5\% fact-checking accuracy with 39.5\% RAcc., while TF-IDF only reached 68.5\% fact-checking accuracy with 11.0\% RAcc. This performance gap stems from dense retrievers' capacity to identify adversarial syntactic paraphrases through embedding space, which TFIDF cannot address effectively. 

\paragraph{\ours is robust to various backbone language models.} As shown in \autoref{tab:factcheckresults}, the accuracy achieved with different LLMs remains consistent. For instance, on \fm, the accuracy difference between \texttt{GPT-4o-mini} and \texttt{Llama-3.1-70B} when using contriver is only 1.5\%. Also, shown in \autoref{tab:whole_factcheckresults}, \ours maintain effectiveness with smaller language models. 
These results indicate that our proposed framework is robust and adaptable to a wide range of LLMs.
\paragraph{\ours also shows feasibility in non-adversarial settings.} Despite not our main focus, we also present results on \wice, a complex fact-checking dataset. 
As shown in \autoref{tab:factcheckresults}, \ours outperforms all baseline approaches, which indicates that our framework is generalizable to other fact-checking scenarios as well. For example, with \texttt{GPT-4o-mini} as the backbone LLM, \ours achieves a remarkable 5.9\% improvement in fact-checking accuracy.



\section{\ours Analysis}
In this part, we will present how each component contributes to the overall performance (\S\ref{sec:ablation}) and provide qualitative analysis on the contribution of each step as well as sources of error (\S\ref{sec:qualitative}).
\subsection{Ablation Study}
\label{sec:ablation}
\begin{table}[!t]
\centering
\scriptsize
\begin{tabular}{lll}
\toprule
Setting & Contriever & TFIDF \\
\midrule
\textbf{\ours (Llama-3.1-70B)} & \textbf{51.0} & \textbf{34.5} \\
\quad $wo.$ claim editing & 49.5 ({\color{blue}-1.5}) & 28.5 ({\color{blue}-6.0}) \\
\quad $wo.$ claim segmentation & 47.5 ({\color{blue}-3.5}) & 27.5 ({\color{blue}-7.0}) \\
\quad Paraphrase $wo.$ Evidence & 50.5 ({\color{blue}-0.5}) & 26.5 ({\color{blue}-8.0}) \\
\midrule
\textbf{\ours (GPT-4o-mini)} & \textbf{54.0} & \textbf{33.5} \\
\quad $wo.$ claim editing & 53.0 ({\color{blue}-1.0}) & 26.0 ({\color{blue}-7.5}) \\
\quad $wo.$ claim segmentation & 52.5 ({\color{blue}-1.5}) & 32.5 ({\color{blue}-1.0}) \\
\quad Paraphrase $wo.$ Evidence & 51.5 ({\color{blue}-2.5}) & 25.0 ({\color{blue}-8.5}) \\
\bottomrule
\end{tabular}
\caption{Ablation study results on \fm. We use Retrieval Accuracy (RAcc.) under Top@10 setting. Numbers in blue parentheses indicate performance drop when removing each component from \ours. }
\label{tab:ablationstudy}
\end{table}
In the previous section, we observed that the retrieval improvement mainly helps the fact-checking ability.
We now delve deeper into why \ours effectively addresses the reasoning-intensive retrieval. To this end, we compare \ours with the following ablations on \fm, using \texttt{GPT-4o-mini} and \texttt{Llama-3.1-70B} as backbone LLMs:
(1) \textbf{wo. Claim Editing}, where we remove the editing module and directly find evidence for each sub-claim;
(2) \textbf{wo. Claim Segmentation}, where we remove the claim segmentation and decontextualization module and directly edit the original adversarial claim;
and (3) \textbf{Paraphrase wo. Evidence}, where we ask LLMs to edit the sub-claims without using retrieved evidence as hints. 

According to \autoref{tab:ablationstudy},\footnote{Complete results in appendix (\autoref{tab:full_ablationstudy}).} our ablation study reveals that all components significantly contribute to the overall performance of \ours, as the removal of any component leads to a notable reduction in performance. Interestingly, while the impact of each component varies, editing grounded on retrieved evidence proves particularly important. 
\paragraph{\textsc{Contriever} is more robust in evidence retrieval.}
As shown in \autoref{tab:ablationstudy}, while TFIDF exhibits a performance drop of 8.5\%, \textsc{Contriever} experiences only a modest 2.5\% reduction. This difference suggests that \textsc{Contriever} is more robust, as it benefits from flexible low-dimensional representations.

\paragraph{Evidence-guided paraphrasing enables reasoning-intensive retrieval.}
For both models and retrieval methods, without claim editing consistently underperforms \ours. With \texttt{Llama-3.1-70B} and TFIDF, paraphrasing without evidence reduces accuracy from 34.5\% to 28.5\%. In addition, removing the evidence during editing leads to significant performance drops, especially with TFIDF where RAcc decreases from 33.5\% to 26.0\% under \texttt{GPT-4o-mini}. 
This reduction highlights the importance of both claim editing and the need of guidence with evidence. 

\paragraph{Sub-claims enhance effective retrieval.}
Removing the segmentation module leads to a significant drop in accuracy.
For \texttt{Llama3.1-70B} with TFIDF, removing claim segmentation decreases RAcc by 7.0\%, which stems from the segmentation module enabling more focused processing of discrete units without requiring complex reasoning.

\begin{figure*}[!t]
    \centering
    \includegraphics[width = \linewidth]{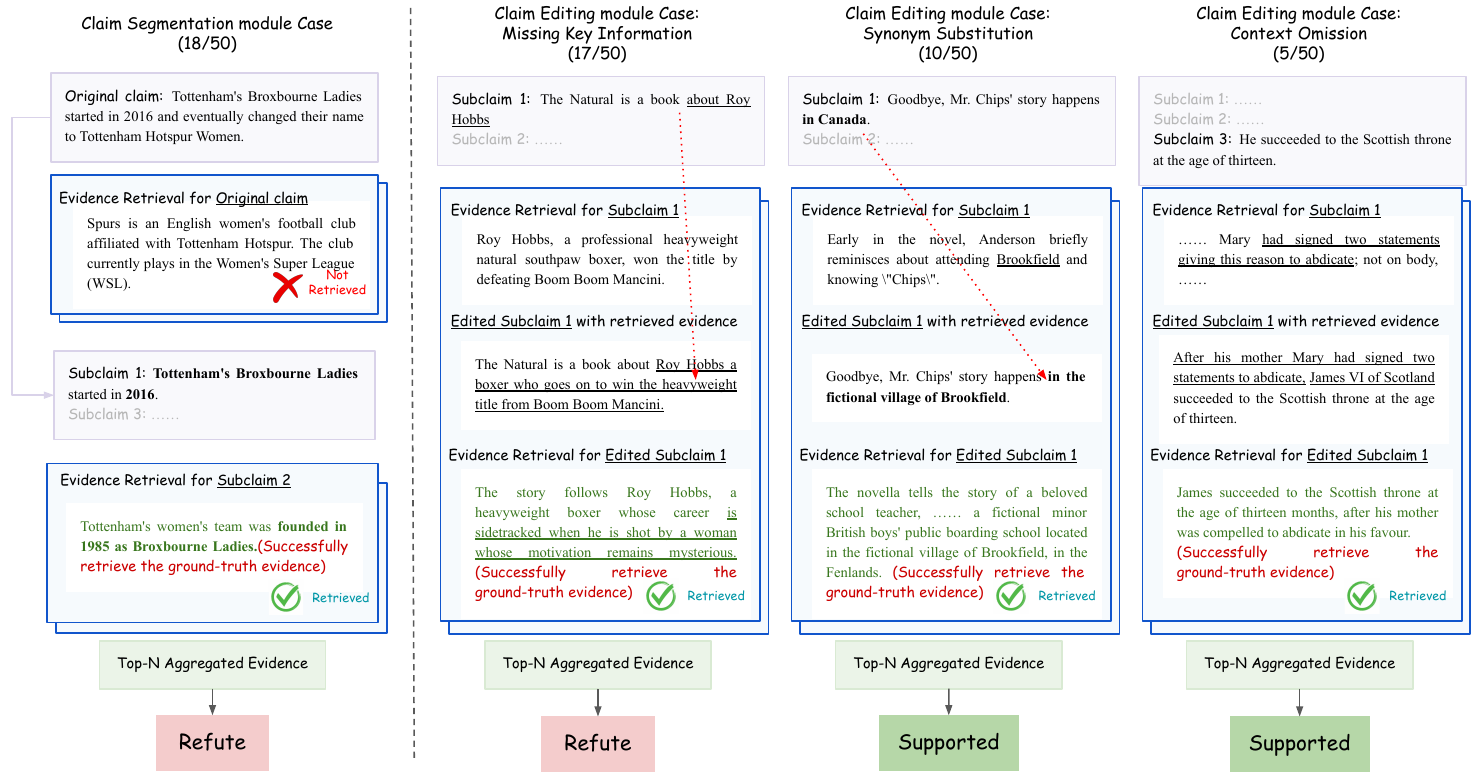}
    \caption{Case studies demonstrating how \ours handles adversarial claims through Claim Segmentation and Claim Editing modules. \textbf{Left}:The original claim is split into sub-claims, enabling successful evidence retrieval. \textbf{Right}: Three cases showing different types of claim editing - missing key information, synonym substitution, and context omission - where the system edits claims to improve retrieval performance. The numbers (\eg 18/50, 17/50) represent the portion of each category over a random sample of 50 correct predictions.
    }
    \label{fig:seg_para_case}
\end{figure*}

\subsection{Qualitative Analysis of \ours}
\label{sec:qualitative}
We next conduct a qualitative analysis to understand how and to what extent each module contributes to the final improvement, using \texttt{GPT-4o-mini} as the LLM backbone and \textsc{Contriver} as the retrieval system.
\paragraph{Claim Segmentation module makes retrieval less distracted.} 
 As shown in the left side of \autoref{fig:seg_para_case}, we observe that the input adversarial claim contains multiple pieces of information, leading to potential confusion in the retrieval process. For instance, ``Tottenham's Broxbourne Ladies started in 2016 and eventually changed their name to Tottenham Hotspur Women.'' prevents the retriever from locating the most important content. With Claim Segmentation component, the retriever focuses on the key fact ``Tottenham's Broxbourne Ladies'' and retrieves the relevant facts.

\paragraph{Claim Editing module makes retrieval easier.}
According to \autoref{fig:seg_para_case}, we find that the Claim Editing module mainly tackles two categories: \textbf{(1) Missing Key Information}, in which some crucial details (\eg entity names) are missing. With the help of retrieved passages as context, LLMs add new information to help the next round of evidence retrieval.
\textbf{(2) Synonym Substitution} that replaces key terms in sentences with their synonyms or more abstract terms to challenge the retrieval systems. As shown in \autoref{fig:seg_para_case}, LLMs address the adversarial challenges by paraphrasing \emph{\underline{``in Canada''}} -> \emph{\underline{``in the fictional village of Brookfield''}}. 
\textbf{(3) Context Omission}, in which background information is deliberately removed to confuse retrieval systems. As shown in \autoref{fig:seg_para_case}, LLMs adds the background information \emph{{``\underline{After his mother Mary had signed two statements} \underline{to abdicate, James VI of Scotland} succeeded to the Scottish throne at the age of thirteen''}}.

\paragraph{Constrained prompt reduces hallucination.}
To mitigate hallucination during editing, we add specific constraints in prompts (\autoref{fig:paraphrase_prompt}), including completing missing  information, replacing statements with evidence-based details, and limiting new information. Our manual analysis of 50 randomly selected samples from \texttt{GPT-4o mini} shows a low hallucination rate of 6.0\%, indicating the effectiveness constrained prompts.

\begin{figure*}[!t]
    \centering
    \includegraphics[width = \linewidth]{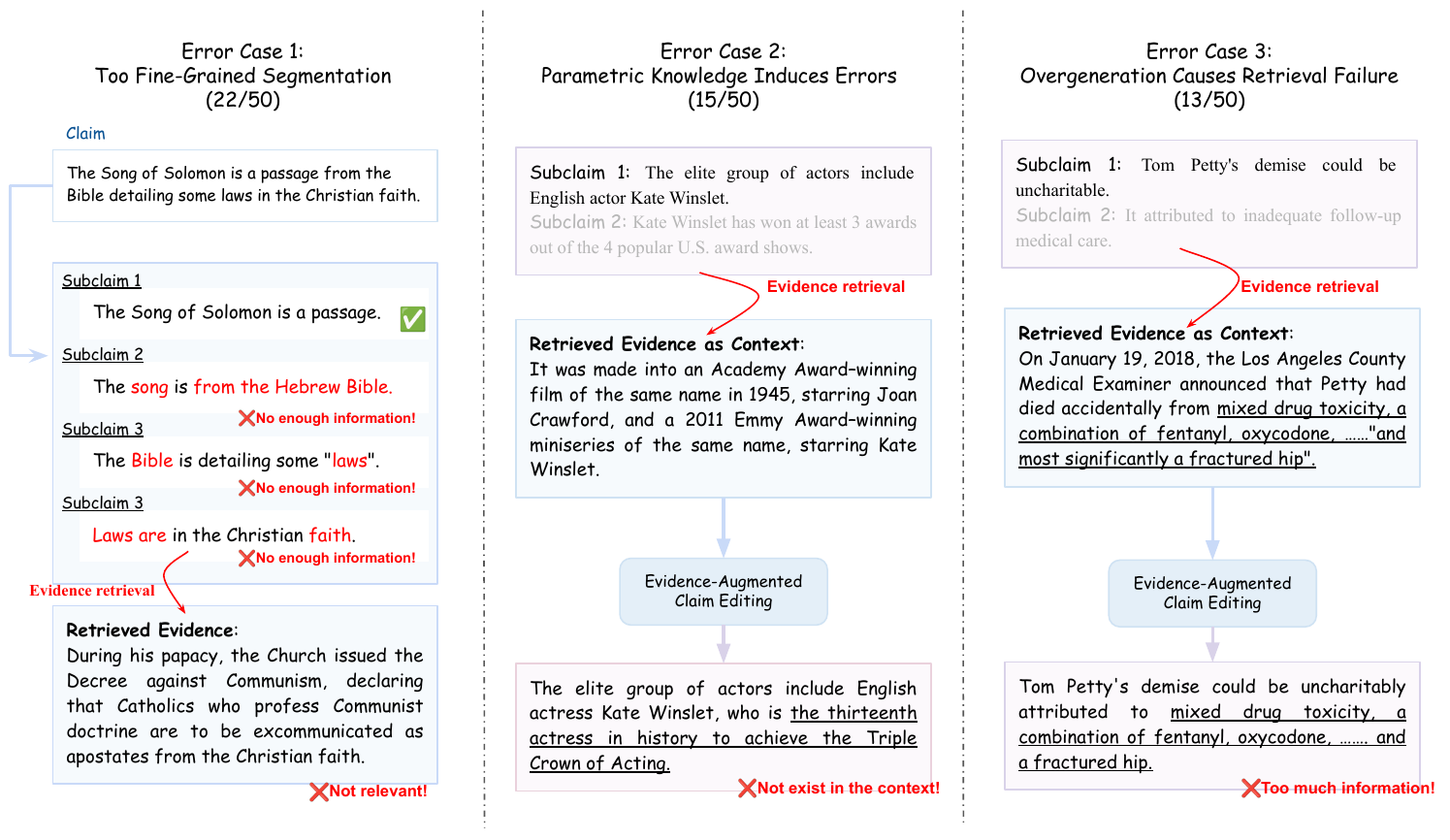}
    \caption{Error analysis showing three main failure cases identified from a random sample of 50 cases: (1) \textbf{Too Fine-Grained Segmentation}, where the Claim Segmentation module breaks claims into overly atomic facts that lack sufficient context, (2) \textbf{Parametric Knowledge-Induced Errors}, where LLM's inherent knowledge introduces unverified information during claim editing, and (3) \textbf{LLM Overgeneration}, where the claim editing module adds additional information from the retrieved evidence. }
    \label{fig:err_case}
\end{figure*}

\subsection{Error Analysis}
We conduct an error analysis To further understand the limitations of \ours. We manually examine 50 sampled instances where the TFIDF method failed within the \texttt{GPT-4o mini} on the \fm dataset.
We identify the following three common mistakes (\autoref{fig:err_case}):  
\textbf{(1) Too Fine-Grained Segmentation:} Atomic facts are segmented with excessive granularity in Claim Segmentation module. As shown in error case 1, the sub-claim does not contain enough useful information, leading to retrieving irrelevant evidence.
\textbf{(2) LLM parametric knowledge leading to hallucination:} Despite using instructions in the prompt, the inherent knowledge of LLMs still introduces extraneous information in claim editing. As shown in error case 2, \texttt{GPT-4o mini} adopts its parametric knowledge as additional information, which results in a deviation in the search direction. 
\textbf{(3) LLM-Overgeneration:} The claim-editing module suffers from generating excessive content. In error case 3,  \texttt{GPT-4o mini} simply adds all information as the edited claim.

\section{Related Work}

\paragraph{Retrieval-Augmented Language Model}
Retrieval-Augmented Language Model~\cite{asai2024reliable, ji2023survey, rawte2024factoidfactualentailmenthallucination, ayala2024reducing,zhao2024docmathevalevaluatingmathreasoning} enhances LLMs by integrating external world knowledge, augmenting LLM's functionality to improve accuracy and factuality (\ie reduce hallucination). Within this paradigm, RARR \cite{gao-etal-2023-rarr}  also include editing modules, but their focus is on the attribution of generated text with retrieved evidence.
Our study is based on RALM, but we extend into a modular framework that better tackles long-form claims with adversarially written segments. 

\paragraph{Reasoning-intensive Retrieval.} Traditional information retrieval tackles the lexical and semantic similarities between query and documents. \citet{su2024bright} introduce a new benchmark about reasoning-intensive retrieval, where retrievers need in-depth reasoning to find relevant documents. We focus on adversarial claims, which also require reasoning capacities to find relevant evidence.

\paragraph{Adversarial Examples in NLP.}
Adversarial examples~\cite{wallace2019universal, goyal2023survey} introduce  perturbations such as entity or sentence paraphrasing that are imperceptible to humans but can mislead NLP systems. Several adversarial datasets have been constructed to evaluate the robustness of NLP systems in various tasks including classification~\cite{garg-ramakrishnan-2020-bae}, natural language inference~\cite{nie-etal-2020-adversarial}, question answering~\cite{jia-liang-2017-adversarial,Bartolo_2020, wallace2019trick} and fact-checking~\cite{eisenschlos-etal-2021-fool}.

\section{Conclusion}
We introduce \ours, a modular framework for verifying adversarial facts. \ours decomposes the complex fact-checking task into three sub-modules: claim segmentation and decontextualization, evidence-augmented claim editing, and evidence aggregation and label prediction. Experimental results on the \fm and \wice datasets illustrate the robust improvements in retrieval efficiency and fact-checking accuracy, surpassing existing claim decomposition based baselines. Our analysis further studies how each module contributes to the improvement and its limitations. We believe this research provides valuable insights into addressing real-world adversarial attacks in fact-checking.
\section*{Limitations}
There are still some limitations in our work:
(1) Due to budget constraints, our evaluation is limited to the test sets of \fm and \wice. The experimental results on the test set are sufficient to show that our framework improves adversarial fact-checking. Future research can expand the test scope to a larger amount of fact-checking datasets.
(2) Our prompting method for claim segmentation and claim editing is still not perfect. In some cases, these prompting methods fail to generate correct subclaims or paraphrase the sub-claims, leading to incorrect fact-checking results. In this case, future work can explore other prompting methods or fine-tune LLMs to better control the LLM generations.
(3) Our work mainly focuses on reasoning-intensive retrieval over adversarial fact-checking, future work can explore other types of reasoning-intensive retrieval tasks, such as retrieving relevant code for user queries. (4) Due to computing resource limitations, we don't cover state-of-art retrievers like BGE-EN-ICL\cite{li2024makingtextembeddersfewshot}. Future work can explore integrating these advanced retrievers.

\section*{Acknowledgements}
Hongjun Liu and Chen Zhao were supported by Shanghai Frontiers Science Center of Artificial Intelligence and Deep Learning, NYU Shanghai. This work was supported in part through the NYU IT High Performance Computing resources, services, and staff expertise.


\bibliography{anthology,custom, llm}

\appendix

\section{Appendix}

\subsection{Multiple-round Iterations}
\begin{figure*}[!t]
    \centering
    \includegraphics[width = 1.0\linewidth]{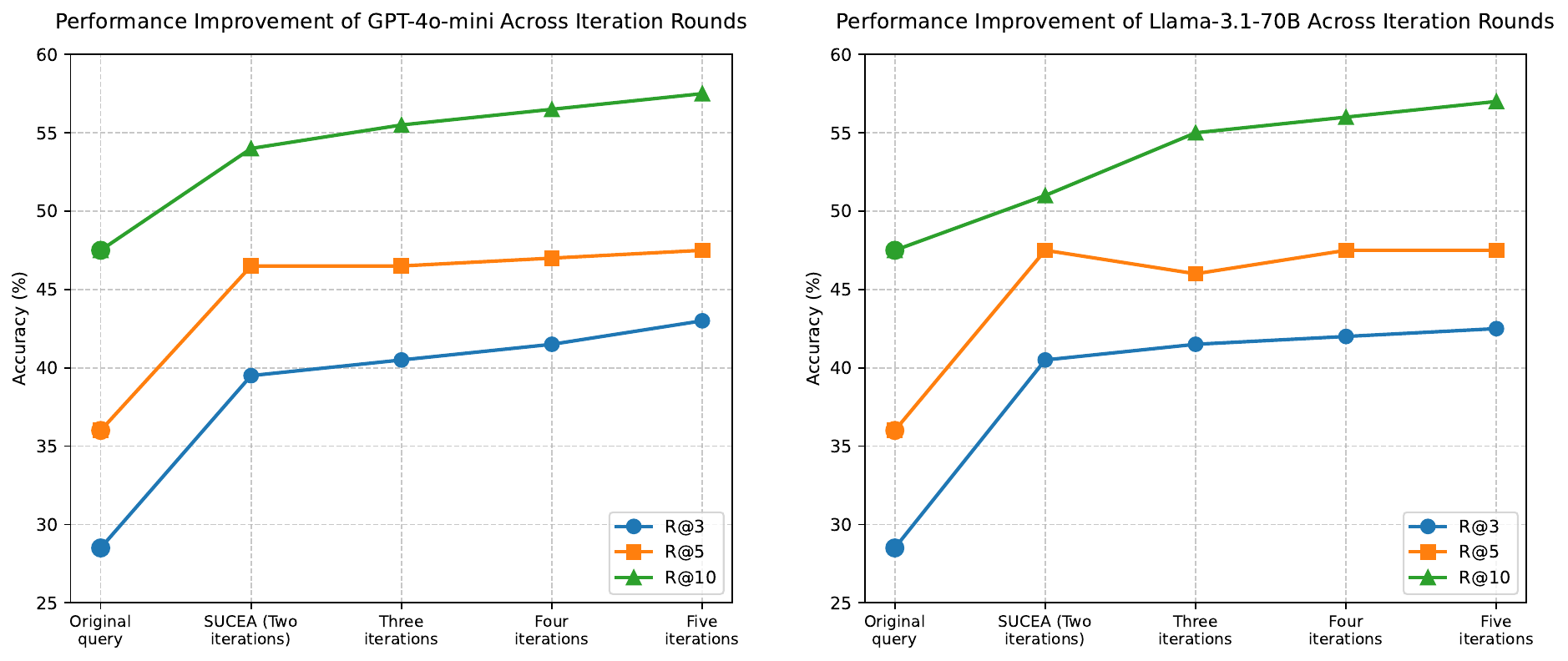}
    \caption{Ablation study results on \fm under multiple-round iteration setting. The performance of both GPT-4o-mini (left) and Llama-3.1-70B (right) improves as the number of iterations increases, with evaluation metrics shown at R@3, R@5, and R@10. Starting from the original query, both models demonstrate significant accuracy gains through \ours's two iterations and subsequent rounds, with the most substantial improvements observed in the early iterations. The R@10 metric consistently achieves the highest performance across all iteration rounds for both models.
    }
    \label{fig:multiround}
\end{figure*}

We conduct an experiment to evaluate the effect on the number of iterations for evidence retrieval and claim editing. As shown in \autoref{fig:multiround}, with more rounds of evidence retrieval and claim editing, the performance steadily improves across all evaluation metrics (R@3, R@5, and R@10). The most substantial gains are observed in the early iterations, particularly after two iterations.

\begin{figure}[h]
\begin{tcolorbox}[colback=black!7.5!white, colframe=black!80!white, title=Claim Segmentation Prompt, fontupper=\footnotesize, fonttitle=\footnotesize]
\texttt{[System Input]}: \vspace{2pt}\\
This is SplitLLM, an intelligent assistant that can split sentences based on their semantic and syntactic structure within the sentence. The following are the sentences you need to split, indicated by number identifier []. I can split them based on their logical units, such as clauses, phrases, or specific grammatical boundaries. Please ensure that each unit retains its original meaning and context for better readability and understanding.\\
\texttt{[0]} \{sen\} \\
Provide the split sentences with each unit separated by a newline.\\
Example: \\
Original: The Natural is a book about Roy Hobbs a natural southpaw boxer who goes on to win the heavyweight title from Boom Boom Mancini.\\
Split: The Natural is a book | about Roy Hobbs | a natural southpaw boxer | who goes on to win the heavyweight | title from Boom Boom Mancini.\\
\newline
\texttt{[User Input]}: \vspace{2pt}\\
Now, proceed with the sentence: \{sen\} The split result of the sentence (only split) is:

\end{tcolorbox}

\caption{Prompt template used for the Claim Segmentation module. The prompt defines the task and expected format, followed by an illustrative example showing how to split a complex claim into logical units.}
\label{fig:seg}
\end{figure}
\begin{figure}[h]
\begin{tcolorbox}[colback=black!7.5!white, colframe=black!80!white, title=Claim Decontextualization Prompt, fontupper=\footnotesize, fonttitle=\footnotesize]
\texttt{[System Input]}: \vspace{2pt}\\
Based on the seperated sentence, if the section misses its subject, complete each split section with proper subject, then form a normal senrence containing enough details. If the the section is a complete sentence, remain the syntax. Here is an example for this split job:\\
  <user> Given sentence: The film High Noon subverts gender norms of the time | by having the woman | rescue the man.\\
  <response> The film High Noon subverts gender norms of the time. | High Noon unfolds by having the woman character. | The woman rescue the man in High Noon. \\
\newline
\texttt{[User Input]}: \vspace{2pt}\\
Given Sentence: \{sen\}
\end{tcolorbox}

\caption{Prompt template for the Claim Decontextualization module. The prompt instructs how to transform segmented claims by adding missing subjects and context to ensure each segment becomes a complete, standalone sentence.}
\label{fig:decontext}
\end{figure}
\begin{figure}[h]
\begin{tcolorbox}[colback=black!7.5!white, colframe=black!80!white, title=Evidence-Augmented Claim Editing Prompt, fontupper=\footnotesize, fonttitle=\footnotesize]
You should complete the sentence <\{query\}> with missing name (author, writer, owner, builder, award-winning titles etc.), number (date, year, population, acreage, etc), location and so on, replace the original adversarial point of view with detailed data or evidence (for example, the biggest population -> population 3,854,000; the date is unknown -> possible dates ranging from 1598 to 1608 ) and correct the counterfactual mistakes in sentence, which is done in order to avoid adversarial cases. You should never add too much new additional information to the sentence. Here is an example:\\
Given sentence: \\
<The album was released in 2018.>\\
Expected result: \\
The "Blackpink in Your Area" compilation album was released in 2018.\\
To help you better complete the required task, we provide following knowledge as contexts: \\
<\{kn\}> \\
Now do the required task on the sentence: <{query}> based on the given knowledge.
\newline

\end{tcolorbox}

\caption{Prompt template for the Evidence-Augmented Claim Editing module. The prompt instructs how to enhance claims by adding specific details from evidence, replacing vague statements with precise information, and correcting inaccuracies.}
\label{fig:paraphrase_prompt}
\end{figure}
\begin{figure}[h]
\begin{tcolorbox}[colback=black!7.5!white, colframe=black!80!white, title=LLM as Reranker Prompt, fontupper=\footnotesize, fonttitle=\footnotesize]
This is RankLLM, an intelligent assistant that can rank passages based on their relevance to the query.  \\
The following are \{len(evidences)\} passages, each indicated by number identifier []. I can rank them based on their relevance to query: \{query\} \\
……\\
(Passages) \\
……\\
The search query is: \{query\}  \\
I will rank the \{top\_k\} passages above based on their relevance to the search query. The passages will be listed in descending order using identifiers, and the most relevant passages should be listed first, and the output format should be [] > [] > etc, e.g., [1] > [2] > [] etc.  \\
The ranking results of the \{top\_k\} passages (only identifiers) is: \\
\newline

\end{tcolorbox}

\caption{Prompt template used for LLM-based evidence reranking. The prompt instructs the LLM to sort retrieved passages based on their relevance to the query. Adapted from \cite{sun2023chatgptgoodsearchinvestigating}.}
\label{fig:ranker}
\end{figure}

\begin{figure}[h]
\begin{tcolorbox}[colback=black!7.5!white, colframe=black!80!white, title=Entailment Label Prediction Prompt, fontupper=\footnotesize, fonttitle=\footnotesize]
\texttt{[System Input]}: \vspace{2pt}\\
You are a well-informed and expert fact-checker. Here are some example of how to act as a professional fact-checker:\\
Claim Example 1: The Cantos is a poem with most of it written over a 40 plus year time span. \\
Evidences for claim example 1: \\
<Most of it was written between 1915 and 1962, although much of the early work was abandoned and the early cantos, as finally published, date from 1922 onwards.
The Cantos by Ezra Pound is a long, incomplete poem in 116 sections, each of which is a canto.
This thread then runs through the appearance of Kuanon, the Buddhist goddess of mercy, the moon spirit from Hagaromo (a Noh play translated by Pound some 40 years earlier), Sigismondo's lover Ixotta (linked in the text with Aphrodite via a reference to the goddess' birthplace Cythera), a girl painted by Manet and finally Aphrodite herself, rising from the sea on her shell and rescuing Pound/Odysseus from his raft.> \\
Step 1: the evidence indicates that The Cantos is a poem with most of it written over a 40 plus year time span.
Step 2: the evidence also mentions that The Cantos by Ezra Pound is a long, incomplete poem in 116 sections. Step 3: the other pieces of evidence provided This thread then runs through the appearance of Kuanon, the Buddhist goddess of mercy, the moon spirit from Hagaromo (a Noh play translated by Pound some 40 years earlier).based on the evidence provided, the claim that The Cantos is a poem with most of it written over a 40 plus year time span is supported. \\
final rating: supported \\
Claim Example 2: \\
......\\
\newline
\texttt{[User Input]}: \vspace{2pt}\\
Now its' your turn, you are provided with evidences regarding the following claim: \{claim\} \\
Evidences: \\
<\{evidence\}> \\
Based strictly on the main claim, and the evidences provided, you will provide: \\
rating: The rating for claim should be one of "supported" if and only if the Evidences specifically support the claim, "refuted" if and only if the Evidences specifically refutes the claim or "not enough information": if there is not enough information to support or refute the claim appropriately. \\
Is the claim: \{claim\} "supported", "refuted" or "not enough information" according to the available questions and answers? \\
Lets think step by step.

\end{tcolorbox}

\caption{Prompt template for the entailment label prediction. The prompt instructs the LLM to act as an expert fact-checker, providing step-by-step reasoning to classify claims as "supported," "refuted," or "not enough information" based on given evidence.}
\label{fig:fm2_entail}
\end{figure}
\begin{table*}[!t]
\centering
\resizebox{\textwidth}{!}{
\begin{tabular}{lccccccc}
\toprule
\multirow{2}{*}{Backbone LLM} & \multirow{2}{*}{Setting} 
& \multicolumn{2}{c}{Top@3} & \multicolumn{2}{c}{Top@5} & \multicolumn{2}{c}{Top@10} \\
\cmidrule(lr){3-4} \cmidrule(lr){5-6} \cmidrule(lr){7-8}
 &   & Contriever & TFIDF & Contriever & TFIDF & Contriever & TFIDF \\
\midrule
{{\textbf{\fm}}} \\\noalign{\vskip 0.5ex}
\multirow{2}{*}{GPT-4o mini}   &\base  & 64.5 & 53.5 & 65.0 & 59.0 & 67.5 & 61.5  \\
&\claimdecompose & 65.0 {\color{red}(+0.5)} & 55.0 {\color{red}(+1.5)} & 65.0 {\color{red}(+0.0)} & 61.0 {\color{red}(+2.0)} & 65.5 {\color{red}(-2.0)} & 63.0 {\color{red}(+1.5)} \\
&\qabrief & 65.5 {\color{red}(+1.0)} & 57.0 {\color{red}(+3.5)} & 67.0 {\color{red}(+2.0)} & 62.0 {\color{red}(+3.0)} & 67.5 {\color{red}(+0.0)} & 62.5 {\color{red}(+1.0)} \\
&\programfc & 68.5 {\color{red}(+4.0)} & 56.0 {\color{red}(+2.5)} & 65.5 {\color{red}(+0.5)} & 61.0 {\color{red}(+2.0)} & 69.0 {\color{red}(+1.5)} & 63.0 {\color{red}(+1.5)} \\
&\minicheck & 66.0 {\color{red}(+1.5)} & 57.0 {\color{red}(+3.5)} & 66.5 {\color{red}(+1.5)} & 62.5 {\color{red}(+3.5)} & 70.0 {\color{red}(+2.5)} & 64.0 {\color{red}(+2.5)}\\
&\textbf{\ours}  & 69.0 {\color{red}(+4.5)} & 61.5 {\color{red}(+8.0)} & 72.5 {\color{red}(+7.5)} & 67.0 {\color{red}(+8.0)} & 75.0 {\color{red}(+7.5)} & 68.5 {\color{red}(+7.0)}  \\
\hdashline
\multirow{2}{*}{LLama 3.1 70B}  &\base  & 61.0 & 56.5 & 64.5 & 59.0 & 65.5 & 61.5 \\
&\claimdecompose & 62.0 {\color{red}(+1.0)} & 57.0 {\color{red}(+0.5)} & 65.5 {\color{red}(+1.0)} & 61.0 {\color{red}(+2.0)} & 67.5 {\color{red}(+2.0)} & 64.0 {\color{red}(+2.5)} \\
&\qabrief & 66.0 {\color{red}(+5.0)} & 56.5 {\color{red}(+0.0)} & 67.5 {\color{red}(+3.0)} & 62.5 {\color{red}(+3.5)} & 66.0 {\color{red}(+0.5)} & 64.0 {\color{red}(+2.5)} \\
&\programfc & 62.0 {\color{red}(+1.0)} & 58.0 {\color{red}(+1.5)} & 65.0 {\color{red}(+0.5)} & 61.0 {\color{red}(+2.0)} & 66.0 {\color{red}(+0.5)} & 64.5 {\color{red}(+3.0)} \\
&\minicheck &67.5 {\color{red}(+6.5)} & 57.5 {\color{red}(+1.0)} & 68.0 {\color{red}(+3.5)} & 62.0 {\color{red}(+3.0)} & 70.5 {\color{red}(+5.0)} & 64.0 {\color{red}(+2.5)} \\
&\textbf{\ours}  & 71.0 {\color{red}(+10.0)} & 63.5 {\color{red}(+7.0)} & 72.5 {\color{red}(+8.0)} & 67.5 {\color{red}(+8.5)} & 73.5 {\color{red}(+8.0)} & 69.0 {\color{red}(+7.5)}  \\
\hline
{{\textbf{\wice}}} \\\noalign{\vskip 0.5ex}
\multirow{2}{*}{GPT-4o mini}   &\base  & 27.2 & 24.0 & 28.6 & 25.8 & 33.7 & 27.3  \\
&\claimdecompose & 29.1 {\color{red}(+1.9)} & 29.1 {\color{red}(+5.1)} & 30.1 {\color{red}(+1.5)} & 30.4 {\color{red}(+4.6)} & 35.2 {\color{red}(+1.5)} & 31.0 {\color{red}(+3.7)} \\
&\qabrief & 28.0 {\color{red}(+0.8)} & 27.5 {\color{red}(+3.5)} & 30.9 {\color{red}(+2.3)} & 27.8 {\color{red}(+2.0)} & 36.2 {\color{red}(+2.5)} & 29.7 {\color{red}(+2.4)} \\
&\programfc & 33.2 {\color{red}(+6.0)} & 26.8 {\color{red}(+2.8)} & 34.8 {\color{red}(+6.2)} & 28.7 {\color{red}(+2.9)} & 38.4 {\color{red}(+4.7)} & 29.9 {\color{red}(+2.6)} \\
&\minicheck &30.0 {\color{red}(+2.8)} & 27.1 {\color{red}(+3.1)} & 31.3 {\color{red}(+2.7)} & 26.9 {\color{red}(+1.1)} & 33.7 {\color{red}(+0.0)} & 30.8 {\color{red}(+3.5)} \\
&\textbf{\ours}  & 34.1 {\color{red}(+6.9)} & 31.0 {\color{red}(+7.0)} & 37.2 {\color{red}(+8.6)} & 28.1 {\color{red}(+2.3)} & 39.0 {\color{red}(+5.3)} & 33.2 {\color{red}(+5.9)}  \\
\hdashline
\multirow{2}{*}{LLama 3.1 70B}  &\base  & 28.1 & 22.1 & 29.6 & 24.5 & 31.2 & 25.7 \\
&\claimdecompose & 26.3 {\color{red}(-1.8)} & 23.0 {\color{red}(+0.9)} & 30.1 {\color{red}(+0.5)} & 25.4 {\color{red}(+0.9)} & 33.0 {\color{red}(+1.8)} & 27.1 {\color{red}(+1.4)} \\
&\qabrief & 30.4 {\color{red}(+2.3)} & 25.1 {\color{red}(+3.0)} & 32.4 {\color{red}(+2.8)} & 27.0 {\color{red}(+2.5)} & 33.5 {\color{red}(+2.3)} & 28.2 {\color{red}(+2.5)} \\
&\programfc & 34.6 {\color{red}(+6.5)} & 24.7 {\color{red}(+2.6)} & 36.5 {\color{red}(+6.9)} & 26.9 {\color{red}(+2.4)} & 33.9 {\color{red}(+2.7)} & 29.3 {\color{red}(+3.6)} \\
&\minicheck  & 39.1 {\color{red}(+11.0)} & 27.8 {\color{red}(+5.7)} & 34.9 {\color{red}(+5.3)} & 27.9 {\color{red}(+3.4)} & 35.2 {\color{red}(+4.0)} & 30.1 {\color{red}(+4.4)} \\
&\textbf{\ours}  & 38.2 {\color{red}(+10.1)} & 30.3 {\color{red}(+8.2)} & 39.0 {\color{red}(+9.4)} & 31.8 {\color{red}(+7.3)} & 38.7 {\color{red}(+7.5)} & 32.9 {\color{red}(+7.2)}  \\
\bottomrule
\end{tabular}
}
\caption{Comprehensive fact-checking evaluation results comparing \ours with baseline approaches on \fm and \wice datasets. Results are reported as accuracy scores across different retrieval settings (Top@3, Top@5, Top@10). Numbers in red parentheses show absolute performance improvements over the \base baseline.}
\label{tab:whole_factcheckresults}
\end{table*}
\begin{table*}[!t]
\centering
\small
\begin{tabular}{lcccccccc}
\toprule
\multirow{2}{*}{Backbone LLM} & \multirow{2}{*}{Setting} 
& \multicolumn{2}{c}{Top@3} & \multicolumn{2}{c}{Top@5} & \multicolumn{2}{c}{Top@10} \\
\cmidrule(lr){3-4} \cmidrule(lr){5-6} \cmidrule(lr){7-8}
 &   & Contriever & TFIDF & Contriever & TFIDF & Contriever & TFIDF \\
\midrule
{{\textbf{\fm}}} \\\noalign{\vskip 0.5ex}
\multirow{2}{*}{LLama 3.1 8B}   &\base  & 58.5 & 50.5 & 59.5 & 51.5 & 60.5 & 56.0   \\
&\textbf{\ours}  & 60.5 {\color{red}(+2.0)} & 58.0 {\color{red}(+7.5)} & 62.5 {\color{red}(+3.0)} & 58.0 {\color{red}(+6.5)} & 65.0 {\color{red}(+4.5)} & 63.5 {\color{red}(+7.5)}  \\
\multirow{2}{*}{Mistral 7B}  &\base  & 48.0 & 46.0 & 56.5 & 46.5 & 55.0 & 49.0 \\
&\textbf{\ours}  & 59.0 {\color{red}(+11.0)} & 50.0 {\color{red}(+4.0)} & 60.5 {\color{red}(+4.0)} & 51.5 {\color{red}(+5.0)} & 63.0 {\color{red}(+8.0)} & 52.0 {\color{red}(+3.0)} \\
\multirow{2}{*}{Gemma2 9B}  &\base  & 44.5 & 39.5 & 46.0 & 43.5 & 54.0 & 48.5 \\
&\textbf{\ours}  & 52.0 {\color{red}(+7.5)} & 49.0 {\color{red}(+9.5)} & 50.5 {\color{red}(+4.5)} & 52.5 {\color{red}(+9.0)} & 61.0 {\color{red}(+7.0)} & 56.0 {\color{red}(+7.5)} \\
\hline
{{\textbf{\wice}}} \\\noalign{\vskip 0.5ex}
\multirow{2}{*}{LLama 3.1 8B}   &\base  & 29.3 & 28.5 & 31.2 & 29.1 & 39.1 & 31.2   \\
&\textbf{\ours}  & 31.4 {\color{red}(+2.1)} & 30.1 {\color{red}(+1.6)} & 34.6 {\color{red}(+3.4)} & 31.7 {\color{red}(+2.6)} & 42.6 {\color{red}(+3.5)} & 31.8 {\color{red}(+0.6)} \\
\multirow{2}{*}{Mistral 7B}  &\base  & 23.1 & 22.6 & 26.9 & 26.5 & 29.0 & 26.9 \\
&\textbf{\ours}  & 25.6 {\color{red}(+2.5)} & 23.2 {\color{red}(+0.6)} & 31.1 {\color{red}(+4.2)} & 26.8 {\color{red}(+0.3)} & 32.5 {\color{red}(+3.5)} & 28.8 {\color{red}(+1.9)} \\
\multirow{2}{*}{Gemma2 9B}  &\base  & 21.5 & 20.7 & 25.9 & 23.9 & 26.6 & 25.2\\
&\textbf{\ours}  & 24.4 {\color{red}(+2.9)} & 23.3 {\color{red}(+2.6)} & 27.4 {\color{red}(+1.5)} & 25.7 {\color{red}(+1.8)} & 30.3 {\color{red}(+3.7)} & 27.4 {\color{red}(+2.2)} \\
\bottomrule
\end{tabular}
\caption{Fact-checking evaluation results comparing \ours with \base baseline using smaller-scale LLMs on \fm and \wice datasets. Numbers in red parentheses indicate absolute improvements over \base, showing that \ours maintains effectiveness even with smaller language models.}
\label{tab:smallmodelresult}
\end{table*}

\begin{table*}[!t]
\centering
    \resizebox{\linewidth}{!}{
    \begin{tabular}{lcccccccccccccc}
\toprule
 \multirow{4}{*}{Model} & \multirow{4}{*}{Setting}  & \multicolumn{4}{c}{Top@3} & \multicolumn{4}{c}{Top@5} & \multicolumn{4}{c}{Top10} \\
        \cmidrule(lr){3-6} \cmidrule(lr){7-10} \cmidrule(lr){11-14}
        & & \multicolumn{2}{c}{Contriever} & \multicolumn{2}{c}{TFIDF} & \multicolumn{2}{c}{Contriever} & \multicolumn{2}{c}{TFIDF} & \multicolumn{2}{c}{Contriever} & \multicolumn{2}{c}{TFIDF} \\
        \cmidrule(lr){3-4} \cmidrule(lr){5-6} \cmidrule(lr){7-8} \cmidrule(lr){9-10} \cmidrule(lr){11-12} \cmidrule(lr){13-14}
        & & Acc. & Recall & Acc. & Recall & Acc. & Recall & Acc. & Recall & Acc. & Recall & Acc. & Recall \\
\midrule
\multirow{5}{*}{Llama3.1 70B}
        & \ours & 40.5 & 35.7 & 27.5 & 23.3 & 47.5 & 40.7 & 29.5 & 26.4 & 51.0 & 46.5 & 34.5 & 29.5 \\
        & wo. claim editing & 39.0 {\color{blue}(-1.5)} & 32.3 {\color{blue}(-3.4)} & 17.5 {\color{blue}(-10.0)} & 15.1 {\color{blue}(-8.2)} & 46.0 {\color{blue}(-1.5)} & 41.1 {\color{blue}(+0.4)} & 21.5 {\color{blue}(-8.0)} & 18.2 {\color{blue}(-8.2)} & 49.5 {\color{blue}(-1.5)} & 46.8 {\color{blue}(+0.3)} & 28.5 {\color{blue}(-6.0)} & 24.4 {\color{blue}(-5.1)} \\
        & wo. claim segmentation & 35.5 {\color{blue}(-5.0)} & 31.0 {\color{blue}(-4.7)} & 16.5 {\color{blue}(-11.0)} & 13.6 {\color{blue}(-9.7)} & 40.5 {\color{blue}(-7.0)} & 35.2 {\color{blue}(-5.5)} & 19.5 {\color{blue}(-10.0)} & 16.7 {\color{blue}(-9.7)} & 47.5 {\color{blue}(-3.5)} & 42.6 {\color{blue}(-3.9)} & 27.5 {\color{blue}(-7.0)} & 22.5 {\color{blue}(-7.0)} \\
        & Paraphrase wo. Evidence & 34.0 {\color{blue}(-6.5)} & 29.8 {\color{blue}(-5.9)} & 14.0 {\color{blue}(-13.5)} & 11.2 {\color{blue}(-12.1)} & 39.5 {\color{blue}(-8.0)} & 35.7 {\color{blue}(-5.0)} & 20.0 {\color{blue}(-9.5)} & 12.3 {\color{blue}(-14.1)} & 50.5 {\color{blue}(-0.5)} & 44.6 {\color{blue}(-1.9)} & 26.5 {\color{blue}(-8.0)} & 21.3 {\color{blue}(-8.2)} \\
\hdashline
\multirow{5}{*}{GPT-4o mini}
        & \ours & 39.5 & 34.9 & 20.5 & 17.4 & 46.5 & 41.1 & 28.5 & 24.0 & 54.0 & 49.2 & 33.5 & 27.9 \\
        & wo. claim editing & 37.0 {\color{blue}(-2.5)} & 32.6 {\color{blue}(-2.3)} & 14.0 {\color{blue}(-6.5)} & 11.6 {\color{blue}(-5.8)} & 45.0 {\color{blue}(-1.5)} & 40.7 {\color{blue}(-0.4)} & 23.0 {\color{blue}(-5.5)} & 19.8 {\color{blue}(-4.2)} & 53.0 {\color{blue}(-1.0)} & 47.7 {\color{blue}(-1.5)} & 26.0 {\color{blue}(-7.5)} & 21.7 {\color{blue}(-6.2)} \\
        & wo. claim segmentation & 36.0 {\color{blue}(-3.5)} & 31.8 {\color{blue}(-3.1)} & 18.0 {\color{blue}(-2.5)} & 15.1 {\color{blue}(-2.3)} & 46.0 {\color{blue}(-0.5)} & 40.3 {\color{blue}(-0.8)} & 23.5 {\color{blue}(-5.0)} & 19.8 {\color{blue}(-4.2)} & 52.5 {\color{blue}(-1.5)} & 46.9 {\color{blue}(-2.3)} & 32.5 {\color{blue}(-1.0)} & 27.1 {\color{blue}(-0.8)} \\
        & Paraphrase wo. Evidence & 29.0 {\color{blue}(-10.5)} & 24.8 {\color{blue}(-10.1)} & 13.5 {\color{blue}(-7.0)} & 10.9 {\color{blue}(-6.5)} & 39.5 {\color{blue}(-7.0)} & 34.1 {\color{blue}(-7.0)} & 20.0 {\color{blue}(-8.5)} & 16.3 {\color{blue}(-7.7)} & 51.5 {\color{blue}(-2.5)} & 44.6 {\color{blue}(-4.6)} & 25.0 {\color{blue}(-8.5)} & 20.5 {\color{blue}(-7.4)} \\
\bottomrule
\end{tabular}
}
\caption{Ablation study results showing the impact of different components on retrieval performance using \fm test set. Results are reported for both accuracy (Acc.) and recall across different retrieval settings. Numbers in blue parentheses show performance drops when removing each component from \ours.}
\label{tab:full_ablationstudy}
\end{table*}

\end{document}